\documentclass[conference]{IEEEtran}
\IEEEoverridecommandlockouts
\usepackage{cite}
\usepackage{amsmath,amssymb,amsfonts}
\usepackage{algorithmic}
\usepackage{graphicx}
\usepackage{textcomp}

\usepackage{acro}
\usepackage{bm}
\usepackage{cite}
\DeclareAcronym{ac}{
    short = AC ,
    long = Actor-Critic
}

\DeclareAcronym{a2c}{
    short = A2C ,
    long = Advantage Actor-Critic
}

\DeclareAcronym{ar}{
    short = AR ,
    long = Actor-Reflector
}

\DeclareAcronym{rl}{
    short = RL ,
    long = Reinforcement Learning
}

\DeclareAcronym{llm}{
    short = LLM ,
    long = Large Language Model
}

\DeclareAcronym{tti}{
    short = TTI ,
    long = Transmission Time Intervals
}

\DeclareAcronym{prb}{
    short = PRB,
    long = Physical Resource Block,
    short-plural = s,
    long-plural = s
}

\DeclareAcronym{ric}{
  short = RIC ,
  long  = RAN Intelligent Controller
}

\DeclareAcronym{ran}{
  short = RAN ,
  long  = Radio Access Network
}

\DeclareAcronym{sla}{
  short = PQoS violation,
  long  = Packet QoS violation
}
\usepackage{booktabs}
\usepackage{subcaption}
\def\BibTeX{{\rm B\kern-.05em{\sc i\kern-.025em b}\kern-.08em
    T\kern-.1667em\lower.7ex\hbox{E}\kern-.125emX}}
\begin{document}

\title{Reflection-Driven Self-Improving Language Model Agents for Adaptive RAN Slicing\\
% consider if we need agnets at the end. I prefer to keep it, but it looks like the title is a bit long.
% \thanks{The work of Yuanhao Li was supported by the China Scholarship Council (CSC) under Grant No.202308060125.}
}

% \and
% \and
% \and
% \and
% \and
\author{
\IEEEauthorblockN{
Yuanhao Li$^{\ast}$, Haozhe Wang$^{\ast}$, Geyong Min$^{\ast}$, Nektarios Georgalas$^{\dagger}$, Wang Miao$^{\ast}$
}
\IEEEauthorblockA{
$^{\ast}$University of Exeter, United Kingdom, Email: \{yl1118, h.wang3, g.min, wang.miao\}@exeter.ac.uk\\
$^{\dagger}$British Telecom, United Kingdom, Email: nektarios.georgalas@bt.com
}
}

\maketitle

\begin{abstract}
Adaptive control in AI-native networks remains challenging because tasks such as dynamic Radio Access Network (RAN) slicing require persistent trade-off decisions among spectrum efficiency, service quality, and reconfiguration stability. Conventional Reinforcement Learning (RL) compresses these competing objectives into handcrafted scalar rewards, requiring exhaustive tuning for a balanced trade-off. Language model (LM) agents provide a new path toward adaptive control because they can semantically represent complex preference relations, but prompt-based adaptation struggles to retain long-horizon interaction history due to finite context limits. We propose a Reflection-Driven Self-Improvement (ReDSI) framework that replaces scalar rewards with semantic feedback for continuous policy adaptation. This framework is built upon a Small Language Model (SLM) agent for fast decision-making, coupled with a bi-perspective reflection to transform interaction history into preference data, subsequently distilling long-horizon decision experience into the SLM parameters through preference-based fine-tuning. Evaluated on a dynamic RAN slicing task, our method achieves an optimal trade-off across multiple competing objectives, outperforming standard RL baselines and existing LM-based agents by demonstrating substantially greater sample efficiency. These results highlight a viable paradigm of self-improving LM agents for reward-free multi-objective continuous control in AI-native networks.
\end{abstract}


\section{Introduction}

The transition to 6G is driven by applications such as holographic telepresence, the Internet of Everything (IoE), and autonomous vehicular networks, which impose stringent demands on latency, throughput, and scalability \cite{6G_survey}. Meeting these demands requires persistent adaptation, making AI-native architecture a key enabler of future networks \cite{nativeai1}. Unlike traditional ``add-on" AI, AI-native systems embed intelligence directly into the network stack, enabling real-time autonomous control and self-optimization under changing traffic, resources, and user demands.

Within the AI-native paradigm, \ac{rl} is widely viewed as a primary tool for closed-loop autonomy, especially for tasks such as dynamic \ac{ran} slicing in 6G, where an agent must continually allocate resources under competing objectives \cite{flexible, AI-Assisted, rl-book}. A good policy must balance latency, throughput, energy efficiency, fairness, and reconfiguration stability rather than optimize a single target. In standard RL, this trade-off is compressed into a handcrafted scalar reward, so the modeling difficulty appears in practice as reward engineering. In dynamic network settings, such rewards are hard to specify and often require exhaustive reward engineering and trial-and-error manual tuning \cite{flexible, eureka}. This raises a key question: can agents learn effective multi-objective control policies without manually engineered scalar rewards?

\begin{figure*}[tbp]
\centering
\includegraphics[width=\linewidth]{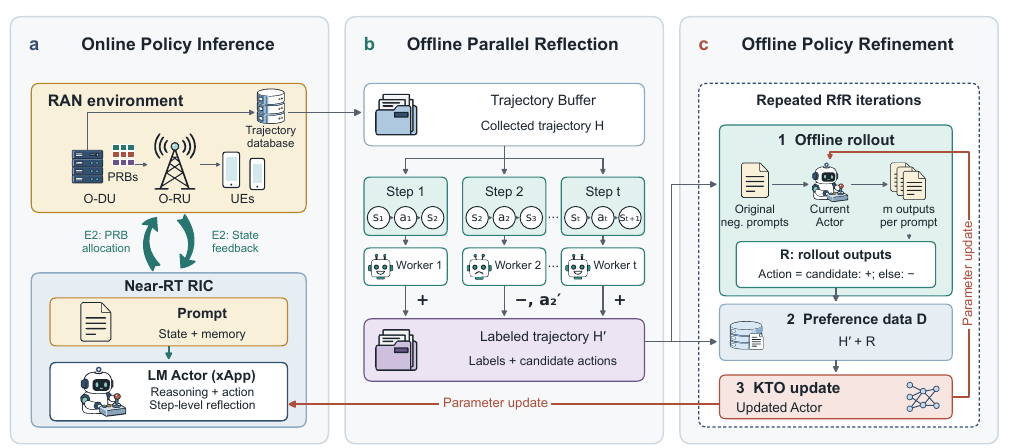}
\caption{Overview of ReDSI. (a) During online policy inference, the Actor allocates PRBs based on state feedback, while its interactions with the RAN environment are recorded as trajectory \(H\). The O-RAN elements illustrate a conceptual deployment mapping. (b) During offline parallel reflection, Reflector workers use the full trajectory \(H\) to label each decision as positive (\(+\)) or negative (\(-\)). They generate a candidate action for each negatively labeled decision, forming the reflection-labeled trajectory \(H'\). (c) During offline policy refinement, each RfR iteration generates rollout outputs \(R\) from original negative-example prompts, combines \(H'\) and \(R\) into preference dataset \(\mathcal{D}\), and updates the Actor through KTO. The same \(H'\) is reused without additional environment interaction. Updated parameters support subsequent RfR iterations and online control.}
\label{fig:system}
\end{figure*}

Language Model (LM) agents offer a promising alternative by utilizing semantic reflection to evaluate past actions, bypassing the need for manually engineered scalar reward signals.
This form of feedback can capture more complex reward patterns than scalar rewards, particularly in multi-objective settings where different goals may conflict. However, the application of current LM architectures to continuous network control reveals critical structural limitations.
Early networking adaptations, such as NetLLM \cite{netllm}, utilize LMs for downstream tasks but rely entirely on offline training with static expert trajectories. Consequently, they fail to support closed-loop continual policy improvement from online environment interactions. 
Recent LM agents, such as Reflexion \cite{reflexion} and ExpeL \cite{expel}, attempt to improve behavior by incorporating textual feedback and past experiences into prompts, either directly or through retrieval from external memory. Although effective for sequential decision-making, these approaches still rely on finite prompt context to exploit accumulated experience. Therefore, these methods remain constrained by finite context windows and long-context degradation \cite{lostinmiddle}, posing challenges for persistent control in AI-native networks.

To address the constraints of prompt-bound memory and the manual tuning required by handcrafted scalar rewards, we propose a Reflection-Driven Self-Improvement (ReDSI) framework deploying a Small Language Model (SLM) agent that internalizes interaction history into model parameters rather than ever-expanding prompts. Through semantic reflection, the framework converts operational experience into preference-based feedback for continuous adaptation. This eliminates scalar reward engineering, enabling sustained autonomous control in dynamic networks.

To realize this ReDSI framework, we make the following key contributions:

\begin{itemize}
    \item We introduce an \ac{ar} framework with bi-perspective reflection, synthesizing step-level and trajectory-level semantic feedback for reward-free policy adaptation.
    
    \item We propose a novel offline distillation pipeline, Refine-from-Reflection (RfR), which converts reflection-labeled trajectories into preference data and distills long-horizon experience into the Actor via KTO \cite{kto}.
    
    \item We evaluate the framework on dynamic RAN slicing and show the strongest overall multi-objective performance with substantially fewer training environment interactions.
\end{itemize}

\section{System Model}

\subsection{AI-RAN Slicing Architecture}

We consider an AI-driven RAN slicing framework for 6G networks, leveraging a state-of-the-art AI-RAN architecture with a central controller deployed on the \ac{ric} to enable adaptive resource allocation across multiple network slices \cite{flexible}. Within this architecture, the system employs an SLM agent to manage the slice resources, which are structured as \acp{prb} in time-frequency grids. The controller dynamically adjusts the inter-slice PRB allocation per decision interval based on monitored performance metrics \cite{flexible}. To ensure isolation, \acp{prb} are strictly partitioned between slices, allowing independent operation within allocated resources.

The framework operates in a closed loop: SLM agents continuously evaluate slice demands and send inter-slice PRB allocation decisions to the controller for the next interval. Intra-slice scheduling is handled by a proportional fair scheduler, while our focus remains on inter-slice control. This design enables adaptive resource management under real-time traffic variation, balancing slice performance requirements and overall spectrum efficiency.

\subsection{Problem Definition}

We formulate the RAN slicing task as a Multi-Objective Optimization Problem (MOOP) with three conflicting goals: maximizing resource utilization, ensuring service quality, and minimizing reconfiguration overhead. Higher utilization improves throughput under bandwidth constraints, service quality requires adaptive allocation to meet diverse QoS needs, and limiting reconfiguration overhead is important because frequent reallocations introduce operational costs \cite{AI-Assisted}. These objectives are \textbf{inherently conflicting}: improving utilization often requires frequent reallocations, which increase reconfiguration overhead and may destabilize service, while prioritizing service quality through over-provisioning can reduce utilization under low demand. Likewise, minimizing reconfiguration overhead through rigid allocations limits adaptation to dynamic traffic. This three-way tension defines the core challenge of efficient RAN slicing.

Spectrum Efficiency (SE) quantifies radio resource utilization. At time step $t$, let $\mathcal{P}_t$ denote the set of packets successfully received by the User Equipments (UEs) in the slice. Then,

\begin{equation}
    SE_t = {\frac{\sum_{p \in \mathcal{P}_t} |p|}{\tau}} / b_t
\label{eq:se}
\end{equation}

\noindent where $|p|$ denotes the size of packet $p$, $\tau$ represents the decision interval, and $b_t$ indicates the allocated bandwidth for the slice at time step $t$.

Service quality is quantified by the cumulative Packet QoS (PQoS) violation $V$. Let $\mathcal{Q}_t$ denote all packets in the slice subject to QoS evaluation at time step $t$, regardless of their reception status. A packet is counted as violating QoS if its metric vector $M_p^t$ exceeds the threshold $\Theta$, indicated by $\chi_p^t = \mathbb{I}(M_p^t \not\models \Theta)$. A timestep is then counted as violating QoS if any packet in \(\mathcal{Q}_t\) triggers such an event. The cumulative \acl{sla} over the measurement window $T_m$ is:

\vspace{-1em}
\begin{align}
    V = \sum_{t=0}^{T_m} v_t = \sum_{t=0}^{T_m} \mathbb{I}\!\left( \exists\, p \in \mathcal{Q}_t : \chi_{p}^{\,t} = 1 \right) %\\
    %\chi_{p}^{\,t} = \mathbb{I}\!\left( M_{p}^{t} \not\models \Theta \right)
    \label{eq:v}
\end{align}

The overhead of resource allocation is measured by the Resource Reconfiguration Times metric $C$, which counts the number of timesteps where the slice bandwidth allocation changes. Let $c_t =\mathbb{I}(b_t \neq b_{t-1})$ denote whether a reconfiguration occurs at timestep $t$. Then,

\vspace{-1em}
\begin{align}
    C &= \sum_{t=0}^{T_m} c_t = \sum_{t=0}^{T_m} \mathbb{I}(b_t \neq b_{t-1})
    \label{eq:c}
\end{align}

These metrics define the finite-horizon MOOP for control policy design over the evaluation window \(T_m\). We seek a policy $\pi \in \Pi$ that maximizes spectrum efficiency while minimizing reconfiguration times and \ac{sla} violations:

\begin{equation}
\begin{aligned}
    &\max_{\pi \in \Pi} \left\{ \mathbb{E}_{\pi}\left[\sum_{t=0}^{T_m} SE_t\right], \right. \\
    &\phantom{\max_{\pi \in \Pi} \{}
    \left. -\mathbb{E}_{\pi}\left[\sum_{t=0}^{T_m}v_t\right],
    -\mathbb{E}_{\pi}\left[\sum_{t=0}^{T_m}c_t\right] \right\}
\end{aligned}
\end{equation}

\section{Proposed ReDSI Framework}

The ReDSI framework comprises online policy inference, offline parallel reflection, and offline policy refinement. We first describe the Actor--Reflector framework, then explain how one reflection-labeled trajectory supports policy refinement.

\subsection{Actor-Reflector Framework}

The design is inspired by the classical \ac{ac} paradigm, in which the Actor selects actions and the Critic evaluates them to guide policy improvement \cite{rl-book}. Our framework follows this decomposition but replaces scalar value estimation with language-based reflection. As shown in Fig. \ref{fig:system}(a), the Actor is an SLM policy \(\pi\) that builds an input sequence \(I_t = \textsc{Prompt}(H_{t-1}, s_t)\) from the current state \(s_t\) and a short interaction history \(H_{t-1}\). The model outputs a structured response \(O_t = \pi(I_t)\), from which we extract a step reflection \(\psi_t\), an action \(a_t\), and a brief action analysis \(\phi_t\). After executing \(a_t\), the environment returns a task-specific metric vector \(M_t\), and the tuple \((s_t, a_t, \psi_t, \phi_t, M_t, I_t, O_t)\) is appended to the trajectory history \(H\).

After each trajectory, the Reflector \(\mathcal{R}\) reviews the full history \(H\) and assesses each decision using the observed outcomes across the episode. It assigns a quality label \(\ell_t \in \{\text{True}, \text{False}\}\) and, when it judges the original decision suboptimal, proposes a potentially better candidate action \(\hat{a}_t\). This converts the raw trajectory into a labeled history \(H'\) (Fig. \ref{fig:system}(b)), which can be processed in parallel across steps and later used for preference-based fine-tuning.

Conceptually, trajectory-level reflection can be represented as an idealized retrospective preference assessment:

\begin{align}
    \hat{a}_t = \arg\max_{a \in \mathcal{A}} \mathbb{E} \left[ \sum_{k=t}^{T} \gamma^{k-t} r_{\text{lang}}(s_k, a_k) \,\middle|\, s_t = s, a_t = a, H \right]
\end{align}

\noindent Here, \(r_{\text{lang}}\) symbolically represents the Reflector's implicit preference over the multi-objective trade-off, rather than an explicit scalar reward supplied by the environment. This expression is a conceptual analogy, not a numerical optimization carried out by the Reflector. In practice, \(\hat{a}_t\) is a candidate proposed using \(H\) and judged potentially better in light of the observed trajectory. The Reflector is analogous to a Critic, but its evaluation is expressed through language-level reasoning rather than an externally engineered numerical reward.

Step-level reflection \((\psi_t, \phi_t)\) provides lightweight prompt memory for online adaptation without changing Actor weights. Trajectory-level reflection uses the complete episode to assess how local actions balance efficiency, service quality, and stability. Together, these complementary views provide real-time guidance and durable supervision.

\subsection{Policy Refinement from Reflection}

Once the Reflector has produced the labeled trajectory \(H'\), ReDSI uses this feedback to refine the Actor through offline preference-based fine-tuning. This policy refinement stage corresponds to Fig. \ref{fig:system}(c) and uses one collected trajectory with its associated reflection labels and candidate actions.

During policy refinement, multiple Refine-from-Reflection (RfR) iterations update the Actor using the same \(H'\). The Reflector assesses the trajectory once, so its labels and candidate actions remain fixed across iterations. Each RfR iteration consists of three steps: rollout with the current Actor, preference dataset construction, and a KTO parameter update.

\noindent \textit{1) Rollout:} For each negative example in \(H'\), we generate \(m\) outputs from the current Actor using the original prompt \(I_t\). Here, rollout means generating LM outputs from stored prompts, without additional environment interaction.

\noindent \textit{2) Preference dataset construction:} Each iteration's preference dataset \(\mathcal{D}\) combines the original Reflector-labeled examples in \(H'\) with the rollout outputs. The original examples are labeled positive for decisions judged effective by the Reflector and negative for those judged suboptimal. For each rollout output, we extract its action and compare it with the fixed candidate \(\hat{a}_t\). Exact action matches are labeled positive, while all other rollout outputs are labeled negative.

\noindent \textit{3) KTO parameter update:} We update the Actor parameters using \(\mathcal{D}\) and the KTO \cite{kto} algorithm. Unlike pairwise preference objectives such as DPO \cite{dpo}, KTO supports unbalanced datasets by directly modeling the absolute preference likelihood of each sample using prospect-theory.

The KTO loss is defined as:

\begin{equation}
\mathcal{L}_{\text{KTO}}(\pi', \pi) = \mathbb{E}_{x, y \sim D} \left[ \lambda_y - v(x, y) \right]
\label{eq:kto_loss}
\end{equation}

\noindent where:

\begin{equation}
v(x, y) =
\begin{cases}
\lambda_{+} \cdot \sigma\left( \beta \cdot (r_\theta(x, y) - z_0) \right), & \text{if } y \sim y_{+} \mid x \\
\lambda_{-} \cdot \sigma\left( \beta \cdot (z_0 - r_\theta(x, y)) \right), & \text{if } y \sim y_{-} \mid x
\end{cases}
\end{equation}

\begin{equation}
r_\theta(x, y) = \log \frac{\pi'(y \mid x)}{\pi(y \mid x)}, \qquad
z_0 = \text{KL}\left( \pi'(\cdot \mid x) \,\|\, \pi(\cdot \mid x) \right)
\end{equation}

Here, \( x \) corresponds to the input prompt \( I \), and \( y \) is the generated output sequence \( O \). The policy \( \pi(y \mid x) \) models the likelihood of generating output \( O \) given prompt \( I \). \(y_{+}\) and \(y_{-}\) denote the corresponding preferred and dispreferred outputs. \( \pi' \) is the policy being optimized, \( \pi \) is the reference Actor fixed within each RfR iteration, and \( \sigma(\cdot) \) is the sigmoid function. The KL term \( z_0 \) constrains policy drift from the reference model. The parameter \( \beta \) controls the sensitivity of the preference signal. The utility function \( v(x, y) \) uses asymmetric scaling coefficients \( \lambda_{+} \) and \( \lambda_{-} \), defined as:

\begin{equation}
\lambda_{+} = \frac{\max(N_{+}, N_{-})}{N_{+}}, \qquad \lambda_{-} = \frac{\max(N_{+}, N_{-})}{N_{-}}
\end{equation}

\noindent where \(N_{+}\) and \(N_{-}\) denote the numbers of positive and negative samples, respectively. 

The first RfR iteration starts from the initial Actor; each later iteration starts from the Actor updated in the preceding iteration. At each iteration's start, the KTO reference is set to this starting Actor version and remains fixed during optimization.

\section{Performance Evaluation}

\subsection{Simulation Environment Settings}

\begin{table}[t]
\centering
\caption{Traffic Model Parameters}
\label{tab:traffic_model}
\begin{tabular}{lll}
\toprule
Metric & \textbf{GBR Traffic} & \textbf{Non-GBR Traffic} \\
\midrule
Active UE count & 20 & 4 \\
Transmission duration & Exp(mean = 15 sec) & Exp(mean = 15 sec) \\
Idle duration & Exp(mean = 15 sec) & Exp(mean = 15 sec) \\
Bit rate & 0.5 Mb/s & 2 Mb/s \\
Packet Size & 512 bytes & 512 bytes \\
QoS Requirement & Delay $<$ 10 ms & Delay $<$ 50 ms \\
\bottomrule
\end{tabular}
\end{table}

\begin{table}[t]
\centering
\caption{Radio Channel Parameters}
\label{tab:radio_channel}
\begin{tabular}{ll} 
\toprule
\textbf{Parameter} & \textbf{Value} \\ 
\midrule
Transmission power & 30 dBm \\ 
Base station antenna gain & 0 dB \\ 
Base station antenna pattern & Antenna Model in 3GPP TR 38.901 \\ 
Noise figure & 5 dB \\ 
Carrier frequency & 2120 MHz \\ 
Propagation model & Urban Propagation Loss Model \\
\bottomrule
\end{tabular}
\end{table}

We evaluate the proposed framework in a custom ns3-based RAN slicing simulator accessed through a Python interface. The task is inter-slice spectrum resource allocation under dynamic traffic. User activity is generated by multiple UEs running independent stochastic on-off application processes within each slice. The on and off durations follow exponential distributions, so the asynchronous switching of many UEs produces stochastic traffic variation over time. The detailed traffic parameters for the two slice types are listed in Table \ref{tab:traffic_model}.

\begin{figure*}[t]
    \centering
    % 第一行
    \subfloat[(a)]{\includegraphics[width=0.24\textwidth]{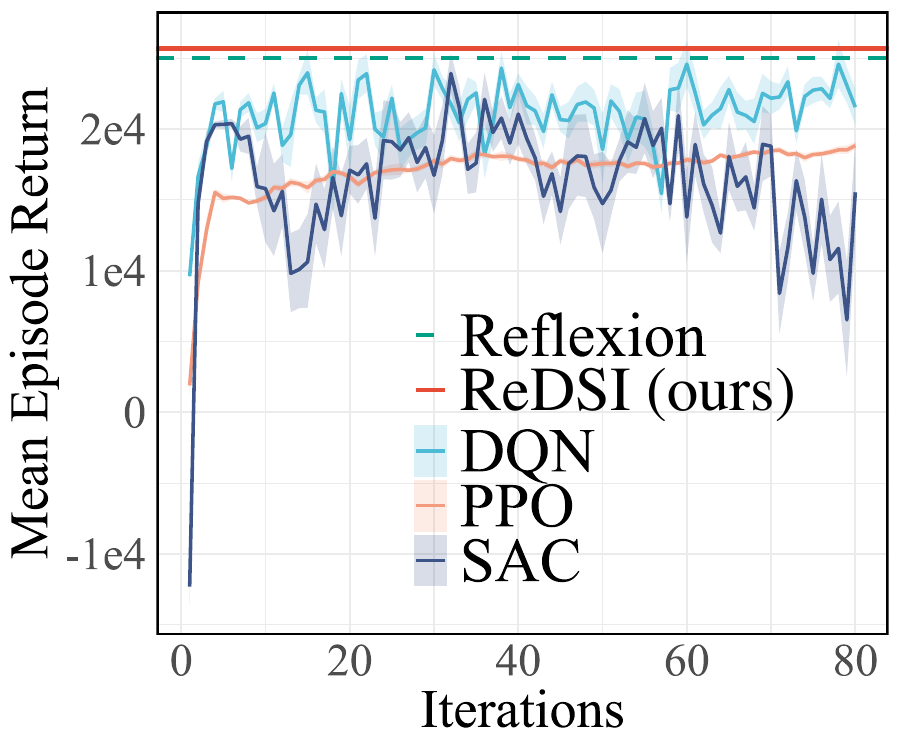}}
    \subfloat[(b)]{\includegraphics[width=0.24\textwidth]{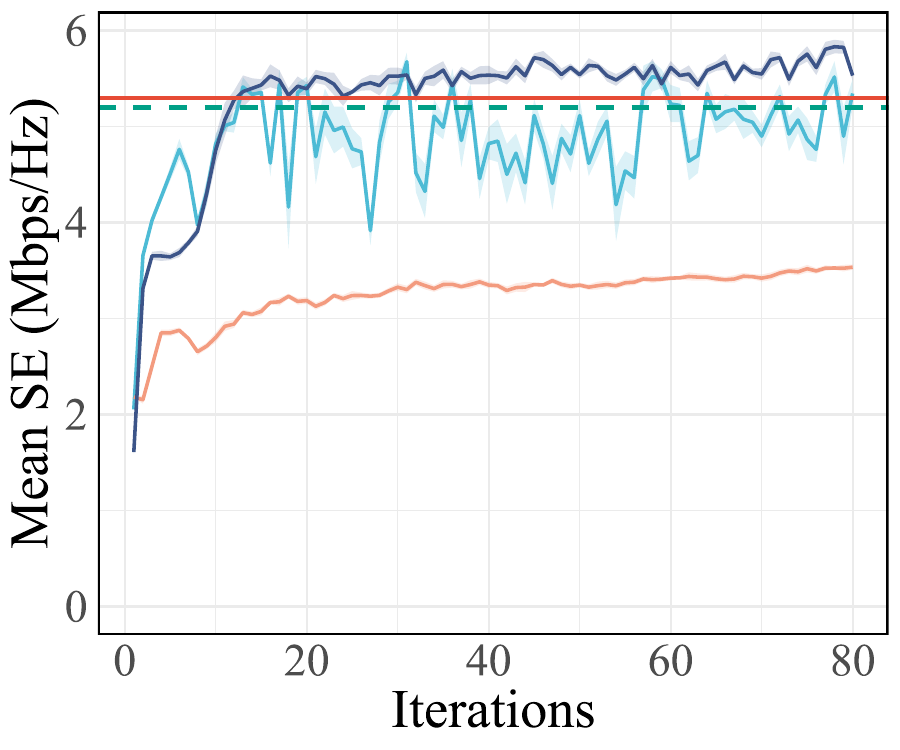}}
    \subfloat[(c)]{\includegraphics[width=0.24\textwidth]{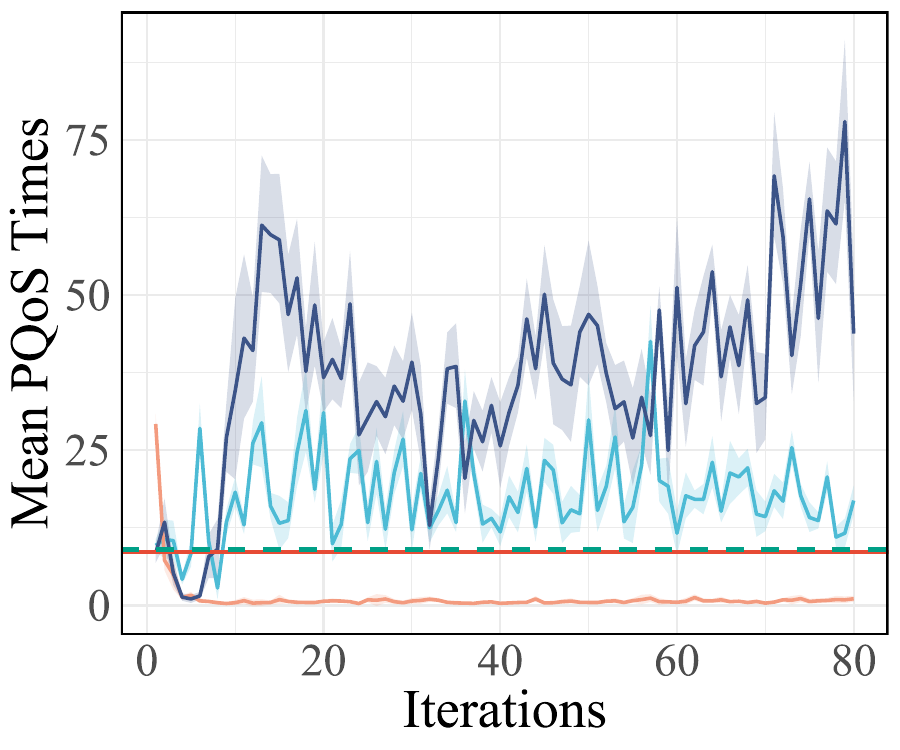}}
    \subfloat[(d)]{\includegraphics[width=0.24\textwidth]{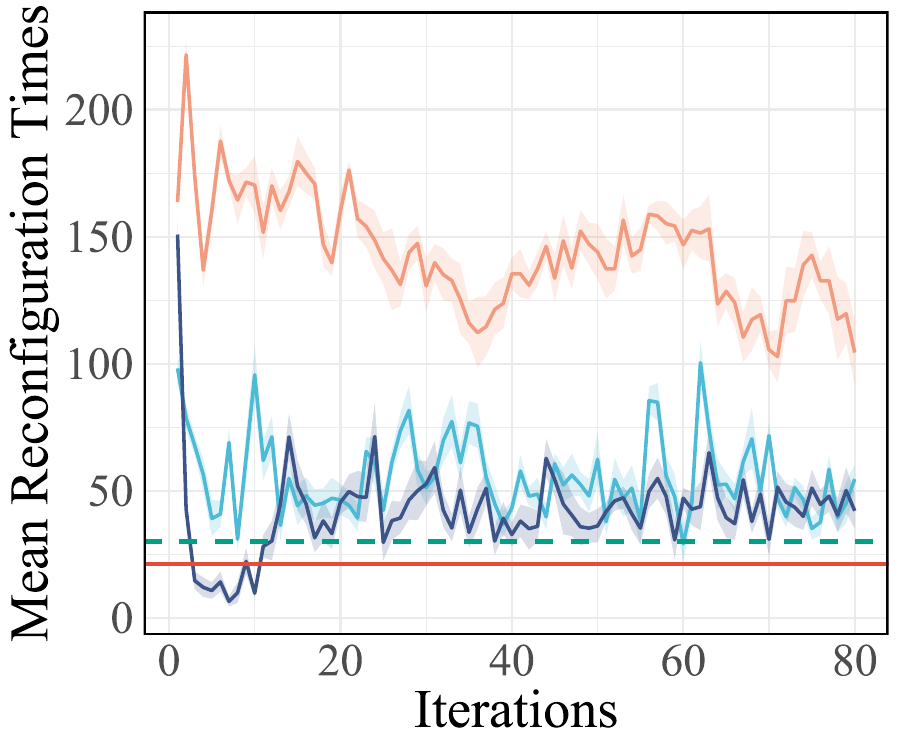}}
    \caption{Performance comparison of RL baselines, Reflexion, and the proposed ReDSI framework across multiple objectives} %spectral efficiency, SLA violations, and reconfiguration times. Our method achieves superior overall balance.}
    \label{fig:exp1}
\end{figure*}

The radio channel follows standard 3GPP-based settings, and the detailed parameters are listed in Table \ref{tab:radio_channel}. To capture realistic channel variability, we additionally introduce frequency-selective fading using pre-generated traces that emulate typical pedestrian and vehicular mobility scenarios.

\subsection{Baseline algorithms}

We evaluate our method against two categories of baselines. First, we implement three RL algorithms: Deep Q-Network (DQN) \cite{dqn}, Soft Actor-Critic (SAC) \cite{sac}, and Proximal Policy Optimization (PPO) \cite{ppo}.

Second, we adapt Reflexion \cite{reflexion} as the primary LM-agent baseline, preserving its tripartite architecture with Qwen3-4B as the Actor and DeepSeek-R1 for trajectory evaluation and self-reflection. For fair comparison, our ReDSI framework uses the same backbone models, so the observed differences are attributable to the learning architecture rather than model capacity.

For all algorithms, the system state at time step t is represented as:

\begin{equation}
s_t = \left[ a_{t-1}, SE_t, \mu_t, \delta_t, \epsilon_t \right]
\end{equation}

\noindent where $a_{t-1}$ is the previous action (current PRB allocation), $SE_t$ is the current spectral efficiency, $\mu_t$ represents the throughput of arriving traffic, $\delta_t$ denotes the increment in queued packet size, and $\epsilon_t$ indicates the size of dropped packets. This compact state representation provides the agent with complete information about current network conditions and resource demands.

The action space represents the number of available PRBs for allocation and is discrete, with the agent determining the optimal resource distribution based on the observed state. The action at each time step directly influences the network's resource utilization and performance metrics.

For RL baselines, we use the following per-slice reward to balance spectral efficiency, allocation stability, and QoS:

\begin{align}
    r_t(s_t,a_t) = \alpha SE_t + (1-c_t)B_{\mathrm{keep}} - v_t P_{\mathrm{QoS}}
\end{align}

\noindent where $\alpha$ weights SE, and $c_t$ and $v_t$ are the binary indicators in \eqref{eq:c} and \eqref{eq:v}, respectively. The term $B_{\mathrm{keep}}(1-c_t)$ rewards unchanged allocations, while $P_{\mathrm{QoS}}$ penalizes QoS violations.\footnote{The experiments below use $\alpha=5$, $B_{\mathrm{keep}}=5$, and $P_{\mathrm{QoS}}=100$.} Episode return sums these rewards over 300 steps and all three slices. For ReDSI and Reflexion, this return is used only for post-hoc evaluation.
\par

\subsection{Experimental Results and Performance Analysis}

\begin{table}[t]
\centering
\caption{Performance comparison (mean $\pm$ standard deviation). `Env. Steps' reports training/adaptation environment steps per run. `Ep. Return' sums per-slice rewards over three slices and 300 steps. The best and second-best results are marked in \textbf{bold} and \underline{underline}, respectively.}
\setlength{\tabcolsep}{3pt}
\begin{tabular}{lccccc}
\toprule
Alg. & Env. Steps & Avg. SE & Reconf. & PQoS & Ep. Return \\
\midrule
ReDSI & 300 & \underline{5.35{\scriptsize $\pm$0.07}} & \textbf{21.09{\scriptsize $\bm{\pm}$6.59}} & \underline{8.56{\scriptsize $\pm$3.76}} & \textbf{25702.2} \\
Reflexion            & 300 & 5.29{\scriptsize $\pm$0.04}         & \underline{29.45{\scriptsize $\pm$9.26}}         & 8.63{\scriptsize $\pm$5.25} & \underline{25314.69}\\
DQN                   & 4.8e5 & 5.21{\scriptsize $\pm$0.26}         & 46.20{\scriptsize $\pm$9.60}         & 15.91{\scriptsize $\pm$4.31} & 22519.1       \\
PPO                   & 4.8e5 & 3.58{\scriptsize $\pm$0.09}         & 119.6{\scriptsize $\pm$21.08}         & \textbf{1.10{\scriptsize $\bm{\pm}$1.04}} & 18532.3 \\
SAC                   & 4.8e5 & \textbf{5.74{\scriptsize $\bm{\pm}$0.09}} & 44.77{\scriptsize $\pm$16.2} & 59.96{\scriptsize $\pm$19.9} & 11704.3 \\
\bottomrule
\end{tabular}
\label{tab:algorithm_comparison}
\end{table}

We report results over 10 random seeds for each RL method and 5 random seeds for each LM method. Evaluation uses 300-step trajectories and reports Mean SE, reconfiguration count, and PQoS-violation timesteps. As illustrated in Fig. \ref{fig:exp1}, we report the learning curves of the RL baselines (SAC, PPO, and DQN) over 80 training rounds, together with the final results of Reflexion and the proposed ReDSI framework under the adopted setup. In addition to these raw metrics, Table \ref{tab:algorithm_comparison} reports the scalarized episode return as a concise measure of overall multi-objective performance.
%DQN underperforms across SE and \ac{sla} compared to other RL baselines.

The RL baselines exhibit clear single-objective optimization biases rather than balanced multi-objective control. From Fig. \ref{fig:exp1}, DQN mainly improves Mean SE, but its reconfiguration curve oscillates around a high level with little net reduction, and its PQoS-violation curve does not improve over training and ends worse than in the early stage. SAC improves Mean SE and partially optimizes reconfiguration, but its PQoS-violation count increases rapidly during training instead of decreasing, indicating that the learned policy shifts toward throughput-oriented behavior at the expense of service quality. PPO behaves in the opposite direction: it achieves the best PQoS-violation result, but its reconfiguration curve declines only slowly and remains high, while its Mean SE stays the lowest among the compared methods. Taken together, these trends show that the RL baselines tend to optimize one or two objectives while failing to maintain balanced control across all three metrics.

In contrast, ReDSI achieves the strongest overall multi-objective performance under the adopted setup while using far fewer training environment interactions. As shown in Table \ref{tab:algorithm_comparison}, ReDSI and Reflexion each use one 300-step trajectory for training/adaptation per run, whereas each RL baseline uses 480,000 training environment steps per run. Despite this gap, ReDSI attains the highest episode return, ranks first or second on every reported metric, and is the only method that remains consistently competitive across all three objectives. Concretely, it achieves the lowest reconfiguration count (21.091), the highest return, the second-highest Mean SE (5.354), and the second-best PQoS-violation count (8.561). PPO still delivers the best PQoS-violation result, and SAC remains the strongest baseline in Mean SE, but both do so by sacrificing the other objectives much more severely. This indicates that ReDSI offers the best integrated balance among efficiency, stability, and service quality, rather than excelling in only one dimension.

Compared with Reflexion, the main gain of ReDSI is a further reduction in reconfiguration count while maintaining similar Mean SE and PQoS-violation performance. Consequently, the advantage of ReDSI is not merely a marginal gain on one metric, but a stronger overall operating point in the three-objective trade-off space. This pattern is consistent with the difference between the two adaptation mechanisms: Reflexion revises behavior through prompt-side history, whereas ReDSI uses RfR to convert trajectory-level reflections into preference data and updates the Actor weights. In this setting, weight-level adaptation appears more effective at retaining useful control patterns across long trajectories.
\begin{figure*}[t]
    \centering
    % Two independently numbered figures share one full-width float.
    \begin{minipage}[t]{0.29\textwidth}\vspace{0pt}
        \centering
        \includegraphics[width=\linewidth]{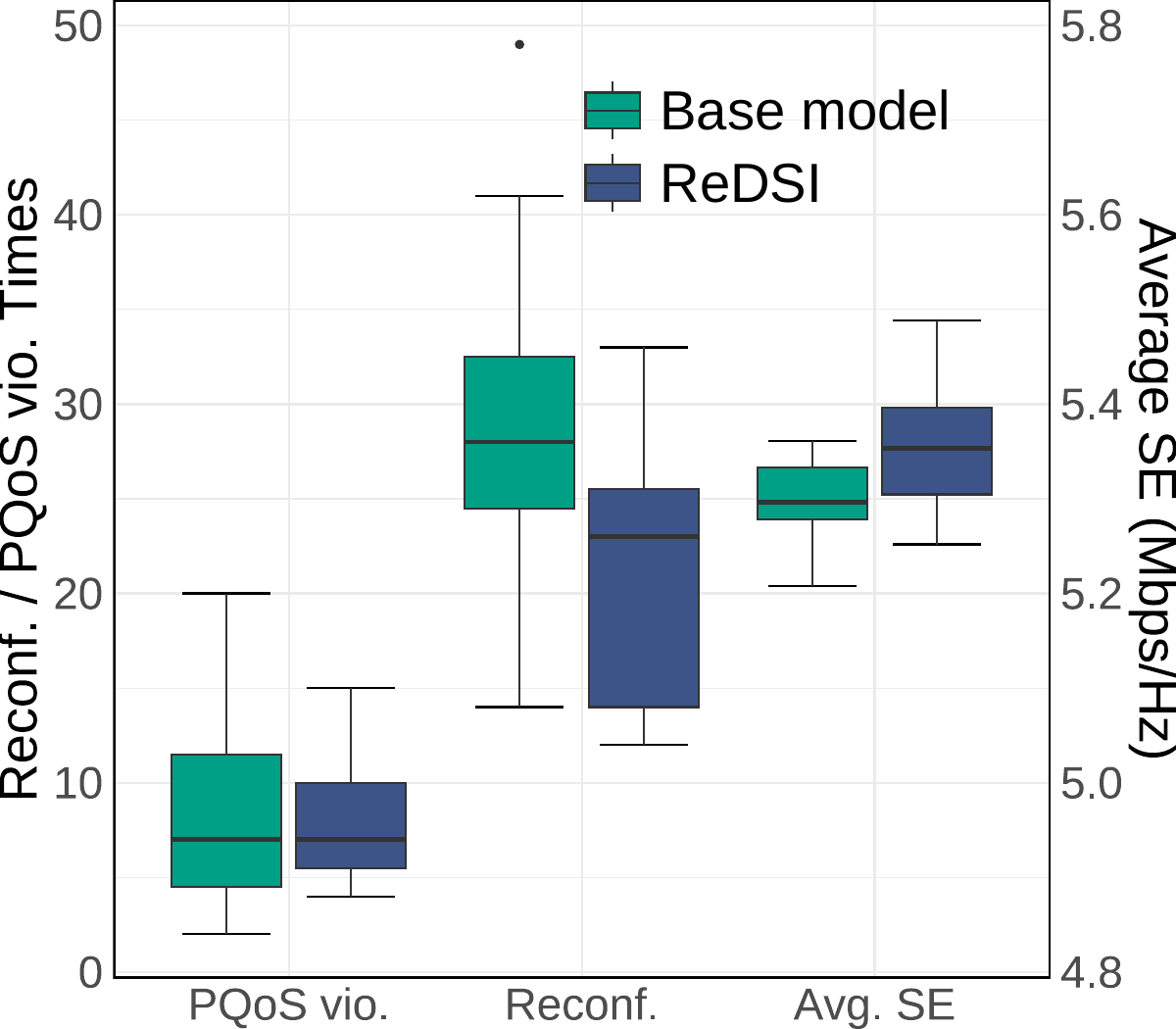}
        % Raise Fig. 3's caption by one line to align both caption bottoms.
        \par\vspace{\dimexpr\baselineskip+4pt\relax}
        \captionsetup{justification=raggedright,singlelinecheck=false}
        \caption{Actor performance before and after policy refinement, showing approximately 33\% fewer reconfigurations.}
        \label{fig:ablation}
    \end{minipage}\hfill
    \begin{minipage}[t]{0.68\textwidth}\vspace{0pt}
        \centering
        % Preserve the internal layout while leaving 3% width between figures.
        \begin{minipage}[t]{0.4927536\linewidth}\vspace{0pt}
            \centering
            \begin{minipage}[t]{0.48\linewidth}
                \subcaptionbox{(a) RfR iteration 1}{\includegraphics[width=\linewidth]{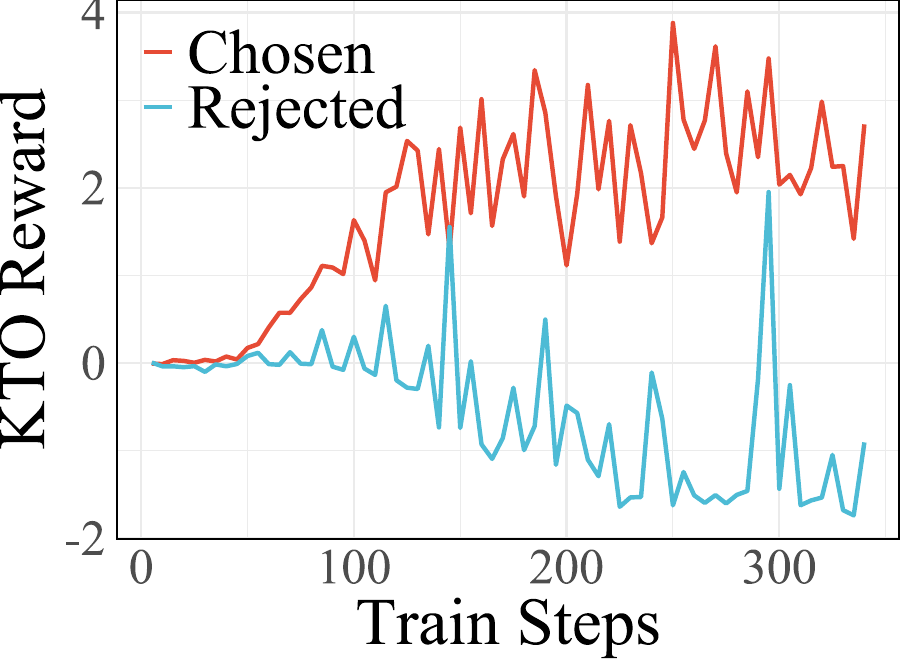}}\\[1mm]
                \subcaptionbox{(e) RfR iteration 5}{\includegraphics[width=\linewidth]{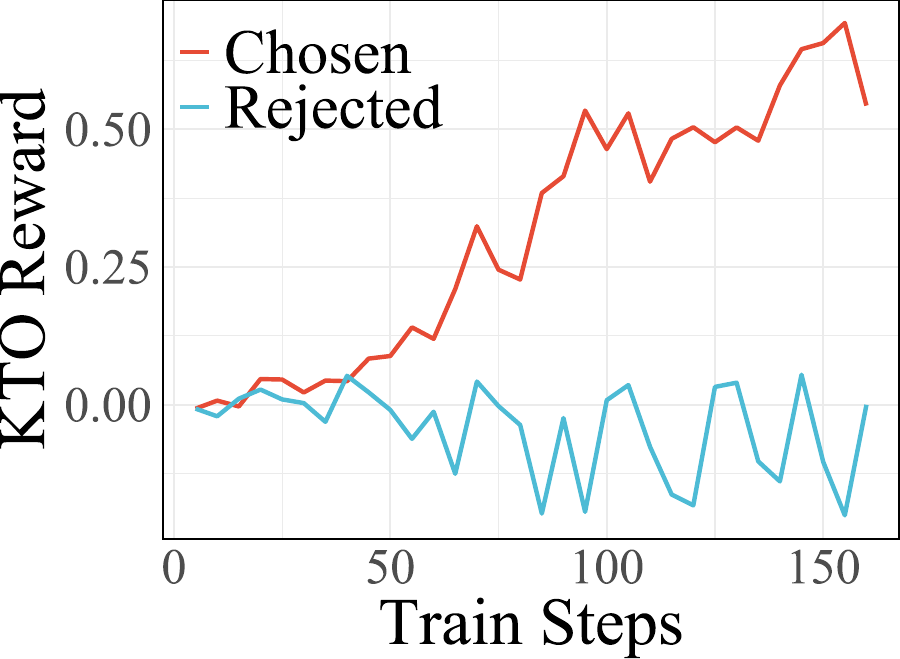}}\\[1mm]
            \end{minipage}
            \begin{minipage}[t]{0.48\linewidth}
                \subcaptionbox{(b) RfR iteration 2}{\includegraphics[width=\linewidth]{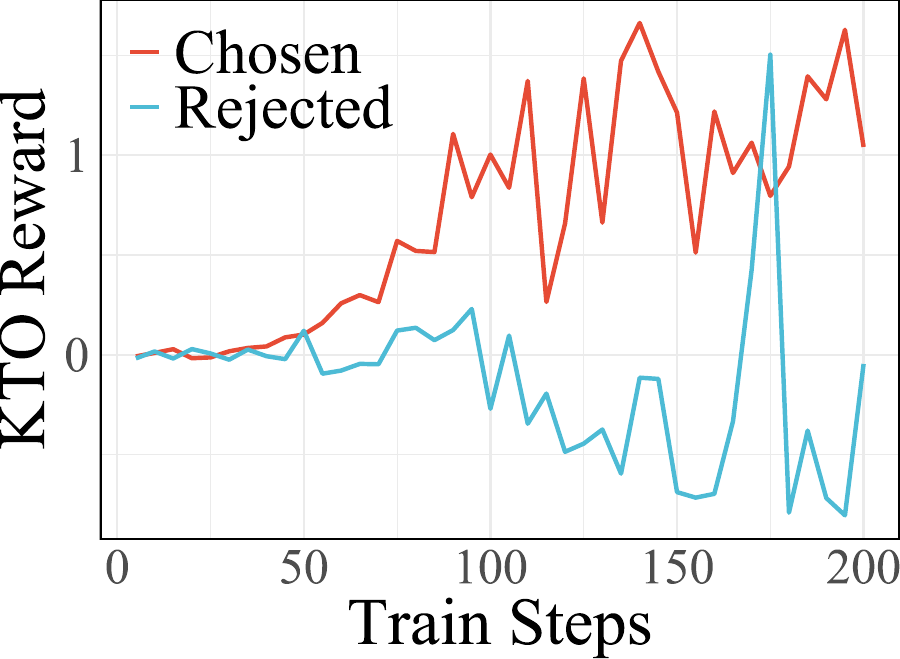}}\\[1mm]
                \subcaptionbox{(f) RfR iteration 6}{\includegraphics[width=\linewidth]{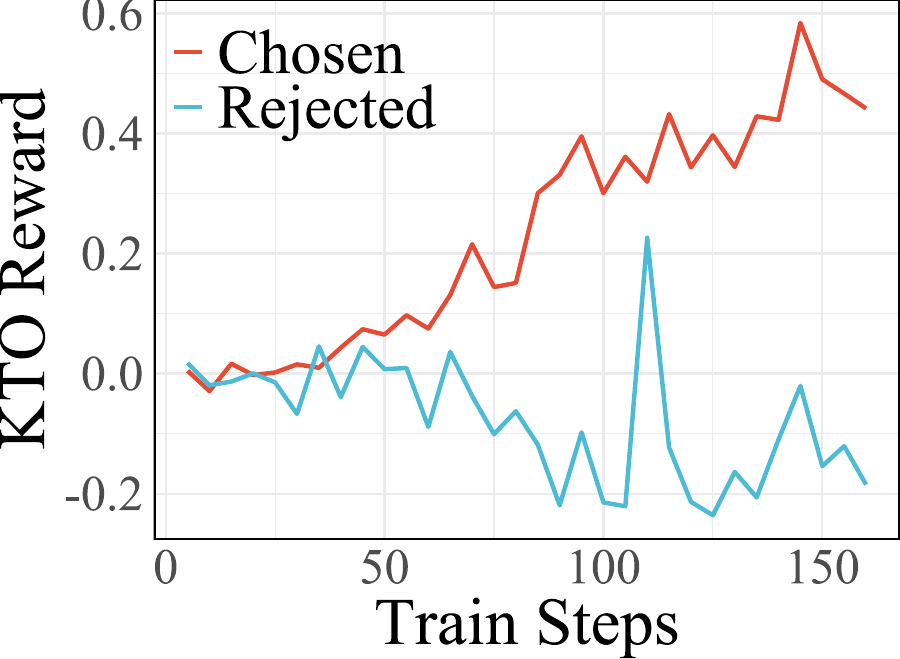}}\\[1mm]
            \end{minipage}
        \end{minipage}
        \begin{minipage}[t]{0.4927536\linewidth}\vspace{0pt}
            \centering
            \begin{minipage}[t]{\linewidth}
                \subcaptionbox{(c) RfR iteration 3}{\includegraphics[width=0.48\linewidth]{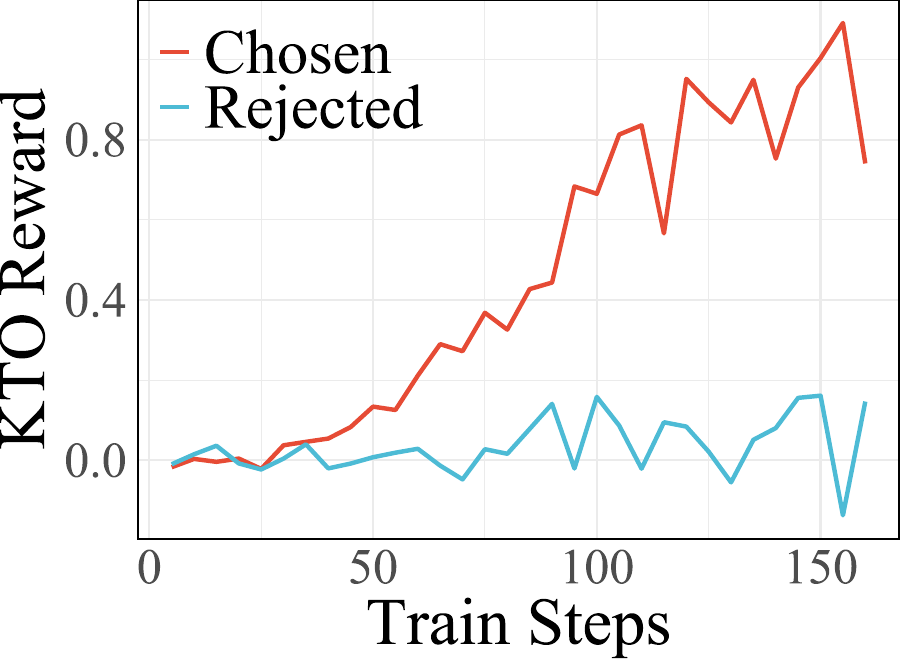}}
                \subcaptionbox{(d) RfR iteration 4}{\includegraphics[width=0.48\linewidth]{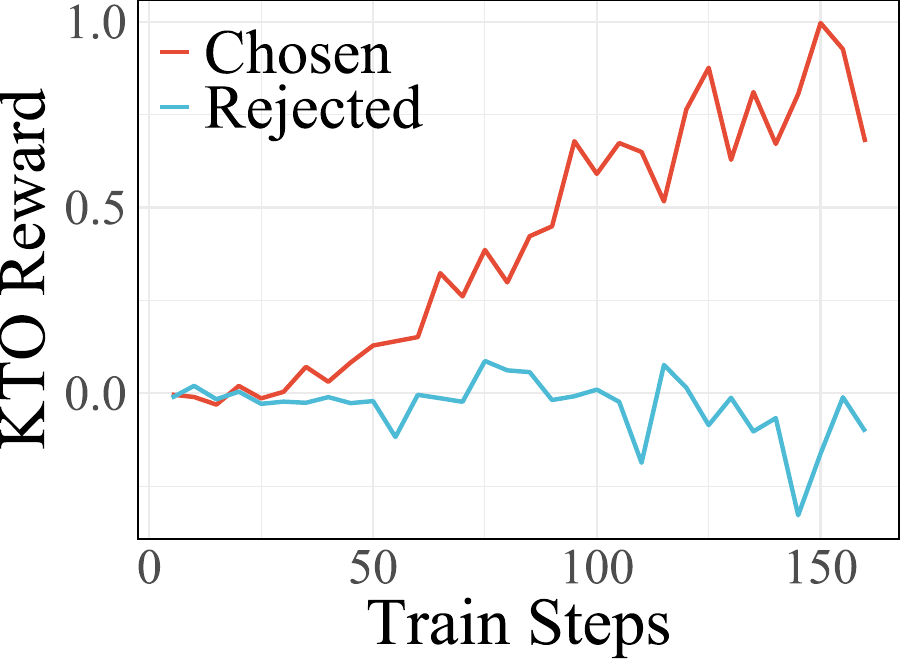}}
            \end{minipage}\\[1mm]
            \subcaptionbox{(g) KTO score convergence}
            {\includegraphics[width=0.96\linewidth]{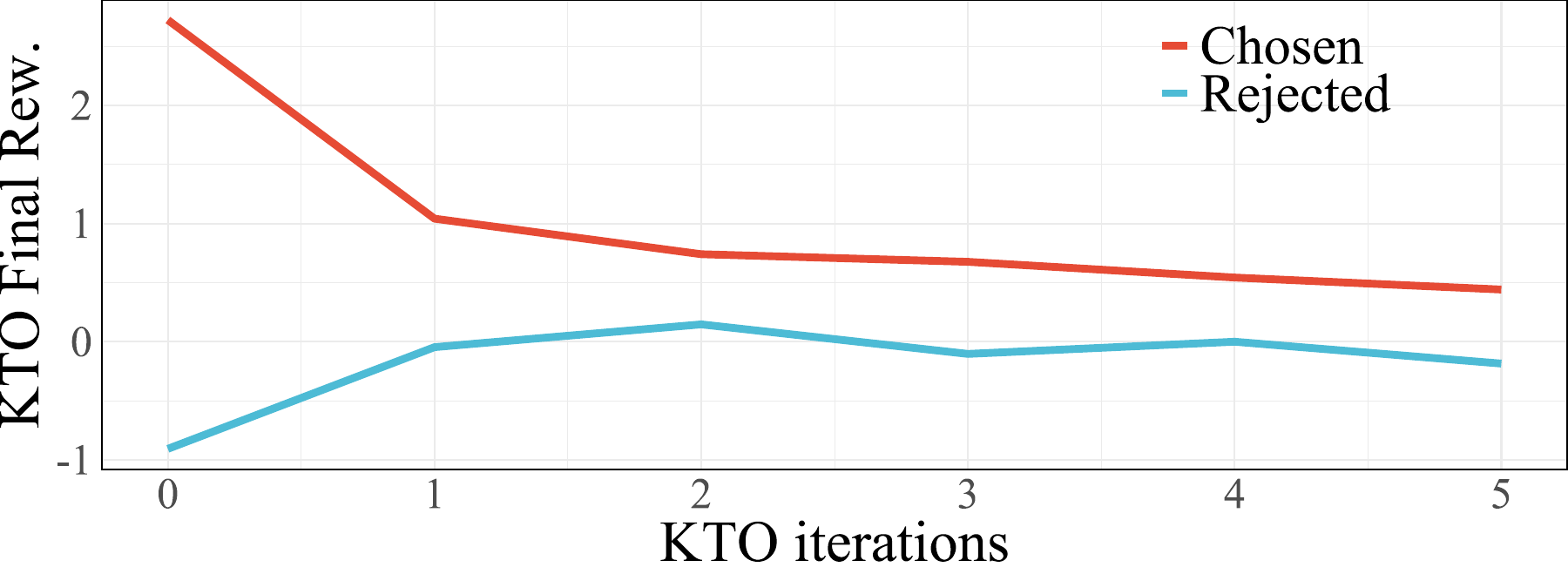}}\\[3mm]
        \end{minipage}
        \caption{KTO training dynamics of ReDSI using one reflection-labeled trajectory. (a)--(f) Chosen and rejected scores across six RfR iterations. (g) Scores gradually approaching zero, indicating diminishing returns from repeated refinement on the same trajectory.}
        \label{fig:kto}
    \end{minipage}
\end{figure*}
\subsection{Ablation Study}

Fig. \ref{fig:ablation} provides an ablation by comparing the pretrained Actor with the same Actor after policy refinement. This refinement stage reuses one reflection-labeled trajectory across six RfR iterations. Each iteration performs rollout, preference dataset construction, and KTO parameter updates. This process reduces the reconfiguration count by approximately 33\% while slightly improving Mean SE and keeping PQoS-violation behavior competitive. This result indicates that the observed gains are not solely inherited from the pretrained backbone, but are associated with reflection-driven policy refinement.

Subplots (a)--(f) of Fig. \ref{fig:kto} show the chosen and rejected KTO scores across the six RfR iterations. These scores are the batch-averaged values of \( \beta r_\theta(x, y_{+}) \) and \( \beta r_\theta(x, y_{-}) \), respectively. The large gap in the first iteration indicates that the initial preference dataset already provides a strong correction signal. As training proceeds, both scores gradually move toward zero, as shown in Fig. \ref{fig:kto}(g), suggesting diminishing returns when repeatedly exploiting the same collected trajectory. After several iterations, the Actor has absorbed most of the useful information from the trajectory.

Overall, these results support sample-efficient adaptation in the studied RAN slicing setting: with minimal additional environment interaction, ReDSI achieves the strongest overall multi-objective performance, as reflected by the highest episode return and first-or-second ranking on every raw metric.

\section{Conclusion}
This paper presented ReDSI for adaptive RAN slicing that balances three competing objectives: spectrum efficiency, service quality, and reconfiguration stability. Step-level reflection guides online decisions, while trajectory-level reflection produces preference data for offline policy refinement. Through RfR and KTO, the SLM Actor internalizes experience beyond prompt memory by repeatedly refining its policy on one labeled trajectory without additional environment interaction.

In the evaluated setting, ReDSI achieves the highest episode return, reflecting overall multi-objective performance, and the fewest reconfigurations. The RL baselines fail to optimize the objectives jointly: PPO suppresses QoS violations but leaves SE lowest and reconfiguration high, whereas SAC and DQN improve SE while QoS violations worsen during training. ReDSI reduces reconfigurations by 28.4\% versus Reflexion with similar SE and PQoS-violation performance under the same adaptation budget. It requires 300 adaptation environment steps versus 480,000 RL training steps per run, supporting sample-efficient multi-objective control without handcrafted scalar rewards.

\bibliographystyle{IEEEtran}
\bibliography{references}

\end{document}